\documentclass[10pt,twocolumn,letterpaper]{article}

\usepackage{icb}
\usepackage{times}
\usepackage{epsfig}
\usepackage{graphicx}
\usepackage{amsmath}
\usepackage{amssymb}
\usepackage{multirow}
\usepackage{float}
\usepackage{lscape}
\usepackage{color}
\usepackage[utf8]{inputenc}
\usepackage{footmisc}

\icbfinalcopy 

\ificbfinal\fi

\begin{document}

\title{FaceQnet: Quality Assessment for Face Recognition based on Deep Learning}

\author{Javier Hernandez-Ortega\\
UAM, Spain\\
{\tt\small javier.hernandezo@uam.es}
\and
Javier Galbally\\
European Commission, JRC, Italy\\
{\tt\small javier.galbally@ec.europa.eu} 
\and
Julian Fierrez\\
UAM, Spain\\
{\tt\small julian.fierrez@uam.es} 
\and
Rudolf Haraksim\\
European Commission, JRC, Italy\\
{\tt\small rudolf.haraksim@ec.europa.eu} 
\and
Laurent Beslay\\
European Commission, JRC, Italy\\
{\tt\small laurent.beslay@ec.europa.eu}
}

\maketitle

\thispagestyle{empty}

\begin{abstract}
   In this paper we develop a Quality Assessment approach for face recognition based on deep learning. The method consists of a Convolutional Neural Network, FaceQnet, that is used to predict the suitability of a specific input image for face recognition purposes. The training of FaceQnet is done using the VGGFace2 database. We employ the BioLab-ICAO framework for labeling the VGGFace2 images with quality information related to their ICAO compliance level. The groundtruth quality labels are obtained using FaceNet to generate comparison scores.
We employ the groundtruth data to fine-tune a ResNet-based CNN, making it capable of returning a numerical quality measure for each input image. Finally, we verify if the FaceQnet scores are suitable to predict the expected performance when employing a specific image for face recognition with a COTS face recognition system.
Several conclusions can be drawn from this work, most notably: 1) we managed to employ an existing ICAO compliance framework and a pretrained CNN to automatically label data with quality information, 2) we trained FaceQnet for quality estimation by fine-tuning a pre-trained face recognition network (ResNet-50), and 3) we have shown that the predictions from FaceQnet are highly correlated with the face recognition accuracy of a state-of-the-art commercial system not used during development. FaceQnet is publicly available in GitHub\footnote{https://github.com/uam-biometrics/FaceQnet}.
\end{abstract}

\begin{figure*}[t!]
\begin{center}
\includegraphics[width=\linewidth]{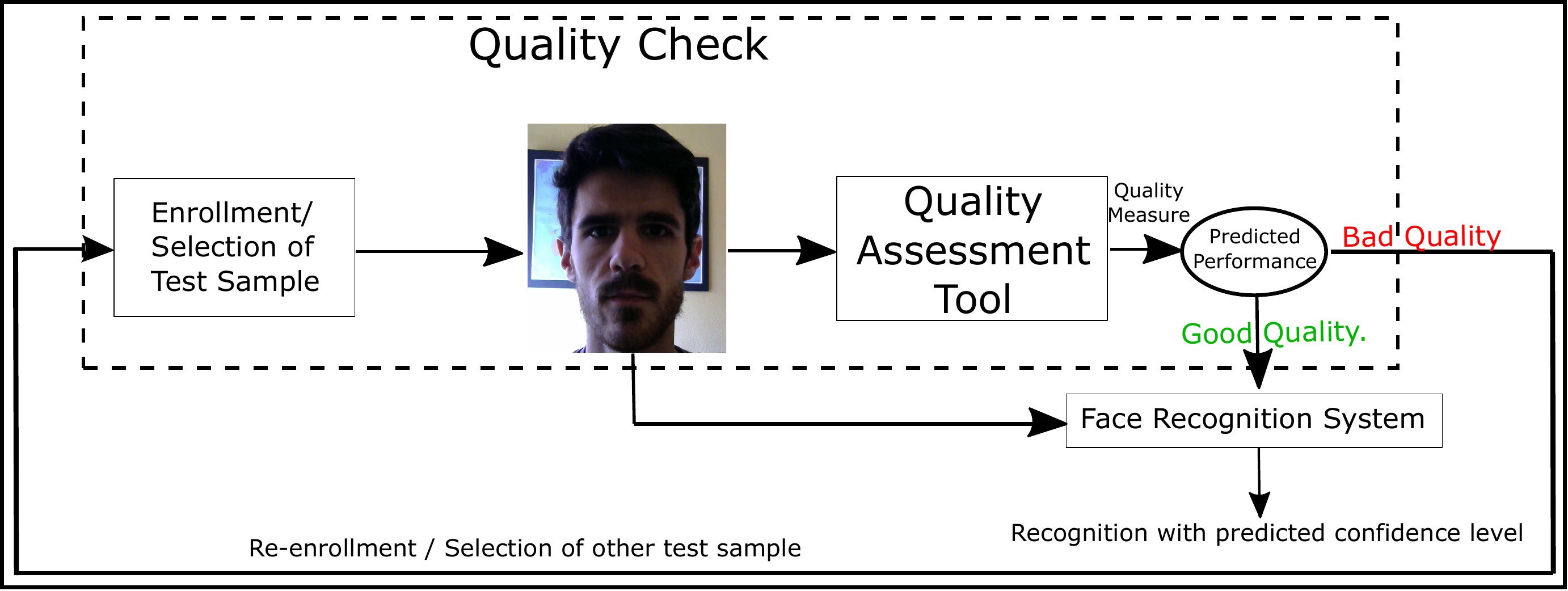}
\end{center}
\caption{\textbf{Example of use of a Quality Assessment (QA) tool} in a face recognition system. The tool can be employed for obtaining feedback at the enrollment stage to have an estimation of the system performance when using the acquired image for face recognition. This way, if the captured image is not of sufficient quality, a re-acquisition strategy can be put in place.}
\label{example_of_use}
\end{figure*}

\section{Introduction}

Face recognition is a biometric characteristic that has received a large amount of attention from researchers in recent years. One of the main properties that differentiates the face from other biometric characteristics, e.g. the fingerprint or the iris, is the possibility of acquisition at a distance, continuously, and non-intrusively. In most scenarios, the accuracy of face recognition systems is the main differential point that makes (or not) its application in the real world possible. 

Traditionally, the use of face for recognition has achieved lower accuracy compared to fingerprint and iris due to the higher variability in the capture conditions. Nowadays, high accuracy levels can be achieved for face recognition in constrained scenarios, in which the users are collaborative, and the acquisition conditions are favorable. However, some of the most relevant applications of face recognition happen under unconstrained conditions \cite{2018_IntelSys_Trends_Proenca}, therefore the ability to deal with variability factors needs to be addressed.  

The performance of face recognition systems is influenced by the variability of samples \cite{2011_QualityBio_FAlonso}. This variability is associated to the image acquisition conditions: illumination, location, background homogeneity, focus, sharpness, etc. Other factors are more related to the properties of the face itself like pose, presence of occlusions, and different expressions. All these factors influence the \textit{quality} of the face samples, which is generally understood as a predictor of the goodness of a given face image to be used for recognition purposes, that is, quality is an estimator of biometric performance.

Developing a monitoring tool to control the quality of face samples, for example in large IT systems where there are multiple acquisition locations for the images (e.g., law-enforcement identification systems), can be very useful. Such a tool could be used to make the enrollment more robust, or for predicting the accuracy level that can be expected for the biometric recognition process. An example of its integration in a face recognition system is depicted in Figure~\ref{example_of_use}. Similar approaches have been used traditionally in fingerprint in order to determine whether the samples present a minimum quality level to be used for recognition purposes. 

\pagebreak 

In this paper we have developed such a tool for face Quality Assessment (QA) based on deep learning.

The rest of this paper is organized as follows: Section \ref{quality_measurement} provides an introduction to quality measures in face biometrics. Section \ref{related_sect} summarizes related
works in face quality assessment. Section \ref{databases} summarizes the datasets used. Sections \ref{proposed} and \ref{evaluation} describe the development and evaluation of our proposed QA system, respectively. Finally, the concluding remarks and the future work are drawn in Section \ref{conclusion_section}.

\section{Introduction to Face Quality Measures}
\label{quality_measurement}

\begin{table*}[t!]
\begin{center}
\resizebox{1\linewidth}{!}{
\begin{tabular}{|c||c||c||c||c||c|}
\hline
Ref & Year & Groundtruth Definition & Type of Input & Features Extracted & Output\\
\hline\hline
\cite{abaza2012quality} & 2012 & Performance-based & Reduced-Reference & {Contrast, bright., focus, sharp. and illum.} & Numerical Score\\
\hline\hline
\cite{ferrara2012face} & 2012 & Human-based & No-Reference & 20 ICAO compliance features & Score from each indv. test\\
\hline\hline
\cite{phillips2013existence} & 2013 & Performance-based & Reduced-Reference & Image, comparator and sensor features & Low/High label\\
\hline\hline
\cite{best2018learning} & 2018 & Human and Performance based &  No-Reference & CNN features & Numerical Score\\
\hline\hline
\textbf{Ours} & 2018 & Performance-based &  No-Reference & CNN features & Numerical Score\\
\hline

\end{tabular}
}
\end{center}
\caption{\textbf{Summary of some representative quality assessment works} in face recognition. The works are classified according to the characteristics introduced in Section \ref{quality_measurement}. The last row shows the method proposed in this paper.}
\label{related_works}
\end{table*}

Essentially, a quality metric in biometrics is a function that takes a biometric sample as its input and returns an estimation of its quality level \cite{2011_QualityBio_FAlonso}. That quality level is usually related to the \emph{Utility} of the sample at hand or, in other words, the expected recognition accuracy when employing the sample in a biometric system.

Other definitions of biometric quality are also possible \cite{2011_QualityBio_FAlonso}: a quality measure can be an indicator of \emph{Character}, i.e., properties of the biometric source before being acquired (e.g., distinctiveness); or a quality measure can also be an indicator of \emph{Fidelity}, i.e., the faithfulness of the acquired biometric sample with respect to the biometric source.

As in most of the quality related works in the literature, for the purpose of the present paper we concentrate on quality measures as predictors of recognition performance. In particular, we will focus on quality measures for face, that can be categorized according to:

\bigskip

\begin{itemize}

\item \textbf{Groundtruth Definition}: One of the main differences between approaches for developing quality measures, is the definition of ``good" and ``bad" quality, i.e., the generation of the groundtruth. Some works employ \textit{human perception} as their groundtruth. Another approach consists in using a \textit{performance-based} groundtruth, which will result in a quality metric that represents the correlation between the input image and the expected face recognition performance of automatic systems.

\item \textbf{Type of Input}: Quality Assessment (QA) systems can be also classified with respect to the amount of information they employ in order to obtain the quality measures. In a \textit{Full-Reference} system (FR), a gallery image with ``good'' quality is supposed to be available. The system compares the features from the probe images with the ones from the high quality reference. In \textit{Reduced-Reference} systems (RR) just partial information of a high quality image is available. \textit{No-Reference} systems (NR) do not have any reference information.

\item \textbf{Features Extracted}: Another distinction can be made among face quality metrics in terms of the features extracted from the images. Face Quality may be influenced by the face itself, e.g. pose, expression, occlusions; by the acquisition sensor, e.g. contrast, resolution, brightness, focus, sharpness, lens distortion; or by the environment, e.g. illumination, background. All these factors can be measured using hand-crafted features (based on traditional digital image processing techniques) or deep-learning features (i.e. features automatically learned by DNNs based on training data).

\item \textbf{Output}: Finally, the output of QA algorithms may differ. Some metrics predict a qualitative \textit{label} for each image in the database to classify them by quality ranges (e.g. low, medium or high quality). Other metrics output a qualitative \textit{decision} declaring whether an image complies with a specific standard or not. Some of the most recent approaches compute a \textit{numerical score} for each input image (e.g., between $0$ and $1$), which serves as a predictor of the expected performance when using that image for face recognition.

\end{itemize}

\begin{figure}[t]
\begin{center}
\includegraphics[width=\linewidth]{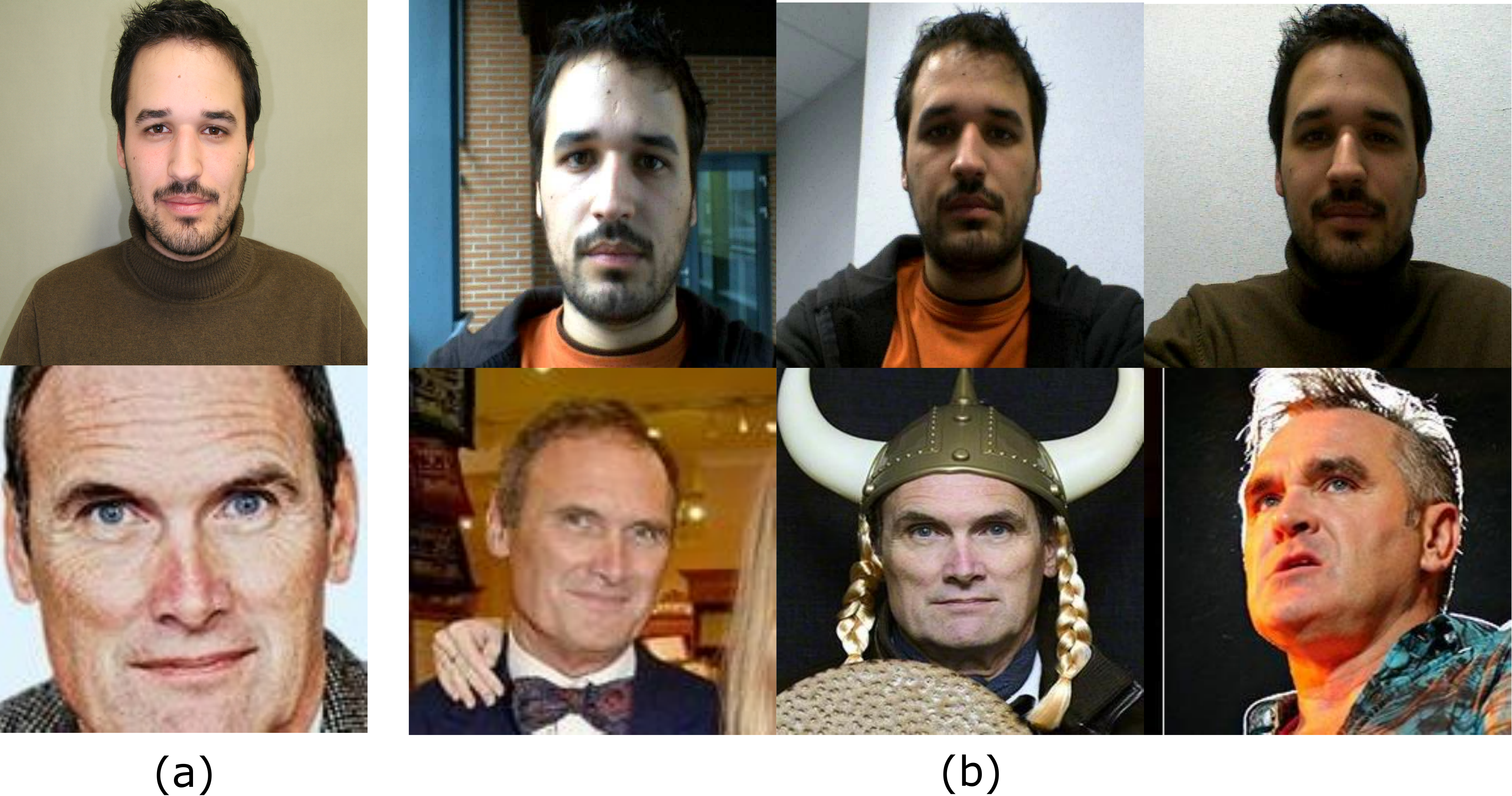}
\end{center}
\caption{\textbf{Examples of face variability} from the BioSecure (top row) and the VGGFace2 (bottom row) databases. (a) shows a gallery high quality image for a random subject. (b) shows examples of low quality images, which suffer from diverse variability factors.}
\label{VGGFace2}
\end{figure}

In this paper we present a Quality Assessment system for face recognition based on \textbf{deep learning}. This system uses a \textbf{performance-based} groundtruth, predicts a \textbf{numerical} quality measure from $0$ to $1$ for the input image \textbf{without} using any other \textbf{reference image}. The quality measure is related to the expected accuracy of the recognition process when using a specific face sample.

\section{Related Works}
\label{related_sect}

\begin{figure*}[t]
\begin{center}
\includegraphics[width=\linewidth]{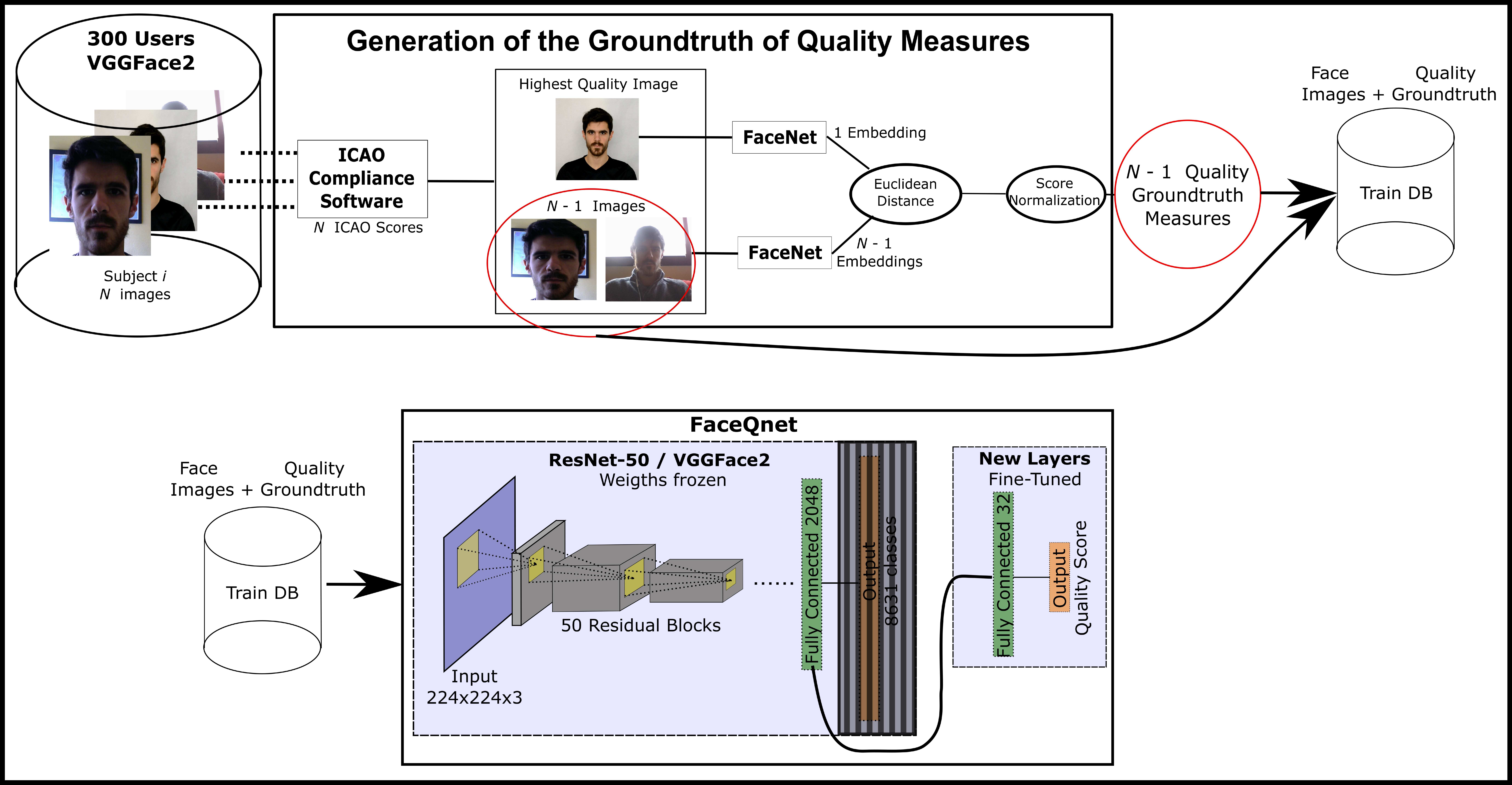}
\end{center}
\caption{\textbf{(Top) Generation of the quality groundtruth}. We selected a subset of $300$ subjects from the VGGFace2 database. We used the ICAO compliance scores for selecting one gallery image for each user. After that, we employed the FaceNet pretrained model for feature extraction \cite{schroff2015facenet}, and we obtained all the mated scores using the Euclidean Distance between the embeddings of the ICAO-compliant gallery images and the rest of the images of the same subject. The normalized comparison scores are used as groundtruth quality measures of the non-ICAO images. We did this process for all the users in the development subset. \textbf{(Bottom) FaceQnet} is based on the ResNet-50 architecture \cite{he2016deep}, but replacing the classification layer with two new ones designed for regression. Using the training set containing face images and their respective groundtruth quality measures, we performed training only of the new layers keeping the weights of the rest frozen.}
\label{general_scheme}
\end{figure*}

In Table \ref{related_works} we include a compilation of some relevant related works in quality assessment for face recognition. Algorithms are classified according to the different characteristics introduced in Section \ref{quality_measurement}. 


The work in \cite{abaza2012quality} is one of the first related to QA for face recognition. The authors proposed a performance-based Face Quality Index (FQI). 

\medskip

They combined individual quality measures extracted from $5$ Digital Image Processing (DIP) features: contrast, brightness, focus, sharpness, and illumination. They defined the global FQI by modeling the distribution of the scores as Gaussian Probability Density Functions (PDFs). Values close to the mean of each PDF mean good quality. The main drawback of this work is that employing Gaussian models may not be a realistic assumption in practical scenarios.

Another approach is described in \cite{ferrara2012face}. The authors presented the BioLab-ICAO framework, an evaluation tool for automatic ICAO compliance checking. The paper defined $30$ different individual tests performed by the framework for each input image. The output consists of a numerical score for each test, going from $0$ to $100$. Those $30$ individual scores were nevertheless not integrated into a final single quality metric.

The work described in \cite{phillips2013existence} used $12$ features divided into three categories: DIP features, sensor-related features from the EXIF headers of the images, and features related to the employed classifiers. The authors extracted conclusions about which of those $12$ features are more relevant to the recognition performance.

Finally, the work in \cite{best2018learning} proposed quality measures related to machine accuracy, and other measures related to human perceived quality. The authors assigned a quality groundtruth to the images from the Labeled Faces in the Wild database according to human perceived quality using the Amazon Mechanical Turk service. The output of the algorithm is a prediction of the machine accuracy and the human perceived quality. They employed a pretrained CNN (VGGFace) for extracting features from the images. Then, they used those features to train an output stage (a SVM). The authors concluded that both measures are correlated to the recognition accuracy, but that human perceived quality is a more accurate predictor.


To date, the work presented in \cite{best2018learning} is probably the most advanced approach to face quality estimation. However, it still presents some drawbacks: 1) a high amount of human effort is required to label their database with human perceived quality; and 2) a manual selection of a high quality image is used for each user to obtain the machine accuracy prediction, this also involves human effort and human bias.


\section{Datasets}
\label{databases}

\subsection{VGGFace2 Database}
In this work we used two disjoint data subsets extracted from the VGGFace2 database \cite{cao2018vggface2}, one for fine-tuning our QA network, i.e. FaceQnet, and the other to evaluate our quality metric with the help of a COTS system for face verification. 

The full database contains $3.31$ million images of $9,131$ different identities, with an average of $362.6$ images for each subject. All the images in the database were obtained from Google Images, and they correspond to well known celebrities such as actors/actresses, politicians, etc. The images have been acquired under unconstrained conditions and present large variations in pose, age, illumination, etc. These variations mean different levels of quality. An example of the images that can be fount in the database can be seen in Figure~\ref{VGGFace2} (bottom).

The creators of the VGGFace2 database also published a CNN based on the ResNet-50 architecture \cite{he2016deep} pretrained with it. 

\medskip 

They showed that they were able to obtain state-of-the-art results when testing their CNN against challenging face recognition benchmarks such as IJB-C \cite{maze2018iarpa}, QUIS-CAMPI \cite{neves2015quis} or PaSC \cite{beveridge2013challenge}.


\subsection{BioSecure Database}
The BioSecure Multimodal Database (BMDBA) \cite{ortega2010multiscenario} consists of $600$ subjects whose biometric samples were acquired in three different scenarios. Images for the first scenario were obtained remotely using a webcam, the second is a more controlled mugshot-type scenario using a high quality camera with homogeneus background, and the third scenario is uncontrolled, captured with mobile cameras both indoors and outdoors.

In this work we have used this database for evaluation purposes. We employed $1,459$ images of $140$ subjects from the second and third scenarios for obtaining quality measures with FaceQnet. An example of the database is shown in Figure~\ref{VGGFace2} (top).



\begin{figure*}[t!]
\begin{center}
\includegraphics[width=0.8\linewidth]{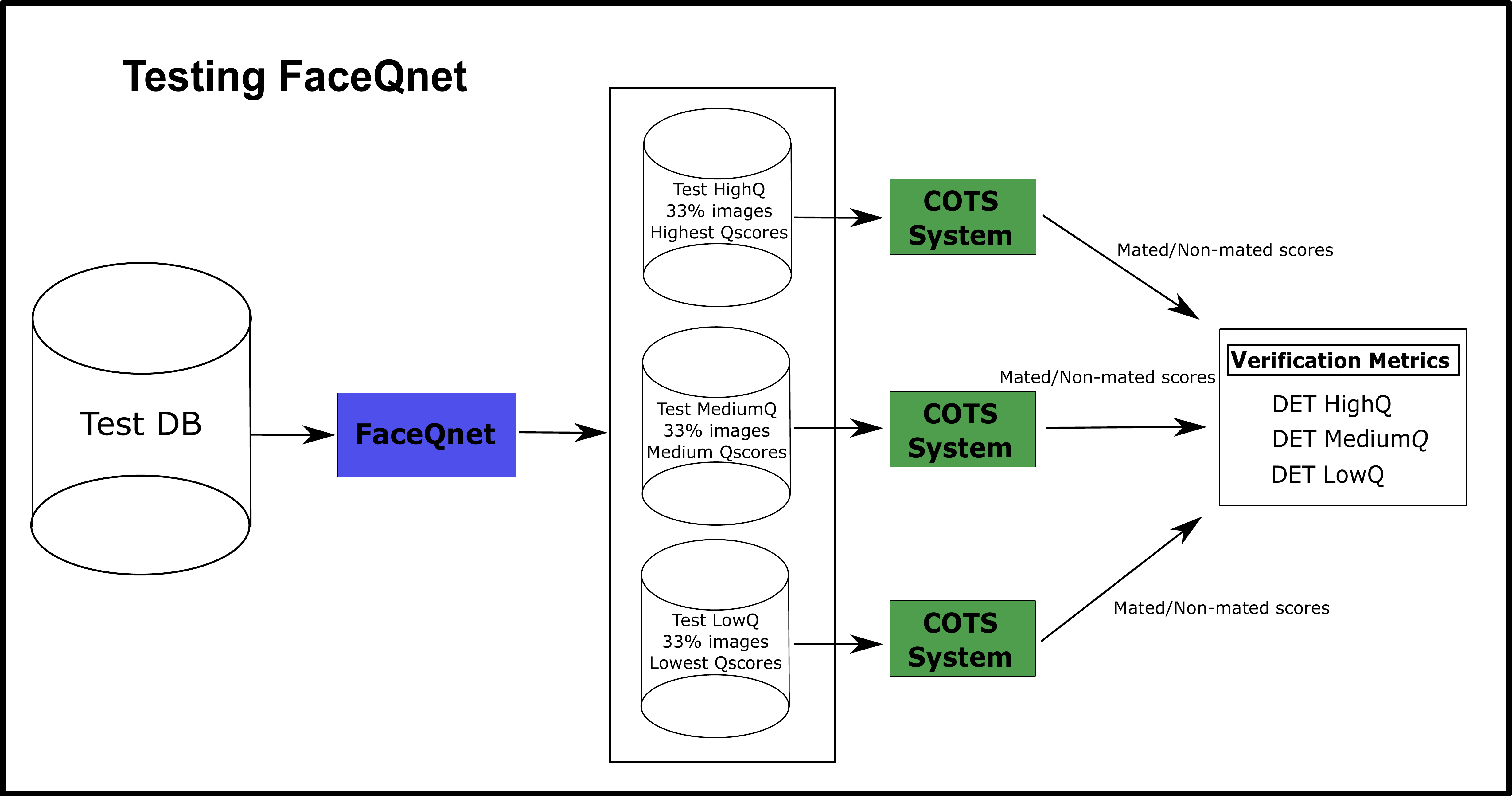}
\end{center}
\caption{\textbf{Experimental scheme for testing FaceQnet}. The quality of all the images in the test database is computed according to FaceQnet. The test database is divided in three sub-databases according to the image quality: 1) Test HighQ, containing the $33$\% of the images with the highest quality; 2) Test MediumQ, containing the $33$\% of the images with medium quality; 3) Test LowQ, containing the $33$\% of the images with the lowest quality. Both mated and non-mated comparison scores are computed using a COTS face recognition system for each of the three test sub-databases. The DET curves of the three sub-databases are extracted and compared. This protocol was followed twice: 1) for a test subset selected from VGGface2 and 2) for a test subset selected from Biosecure.}
\label{validation_scheme}
\end{figure*}

\section{FaceQnet: Development}
\label{proposed}


Our system was designed as an extension of the work presented in \cite{best2018learning}. Our objective is to correlate the quality of an image to its expected accuracy in face recognition. That is, we want to correlate quality measures to comparison scores.

We followed a similar approach to QA than \cite{best2018learning}, but we attempted to tackle its shortcomings. To that end: 1) we did not generated the groundtruth based on human perceived quality, only based on performance quality values; and 2) we employed a third party software to obtain automatically ICAO compliance scores in order to select a high quality gallery image for each subject, thereby avoiding the introduction of bias from human operators; We fine-tuned a pre-trained face recognition network (ResNet-50) to perform the quality prediction.

Biometric quality estimation can be seen as a prediction of biometric accuracy, that is, a regression problem. There are two main steps to be solved in this process, as is shown in Figure \ref{general_scheme}: 1) the generation of the groundtruth database (images + groundtruth quality scores) in order to train the regression model; and 2) the selection of the features extracted from the images to be used within the regression model. Each of the next subsection describes the process followed to address each of these tasks.


\subsection{Generation of the Groundtruth}
\label{groundtruth_gen}


\begin{figure*}[t]
\begin{center}
\includegraphics[width=0.8\textwidth]{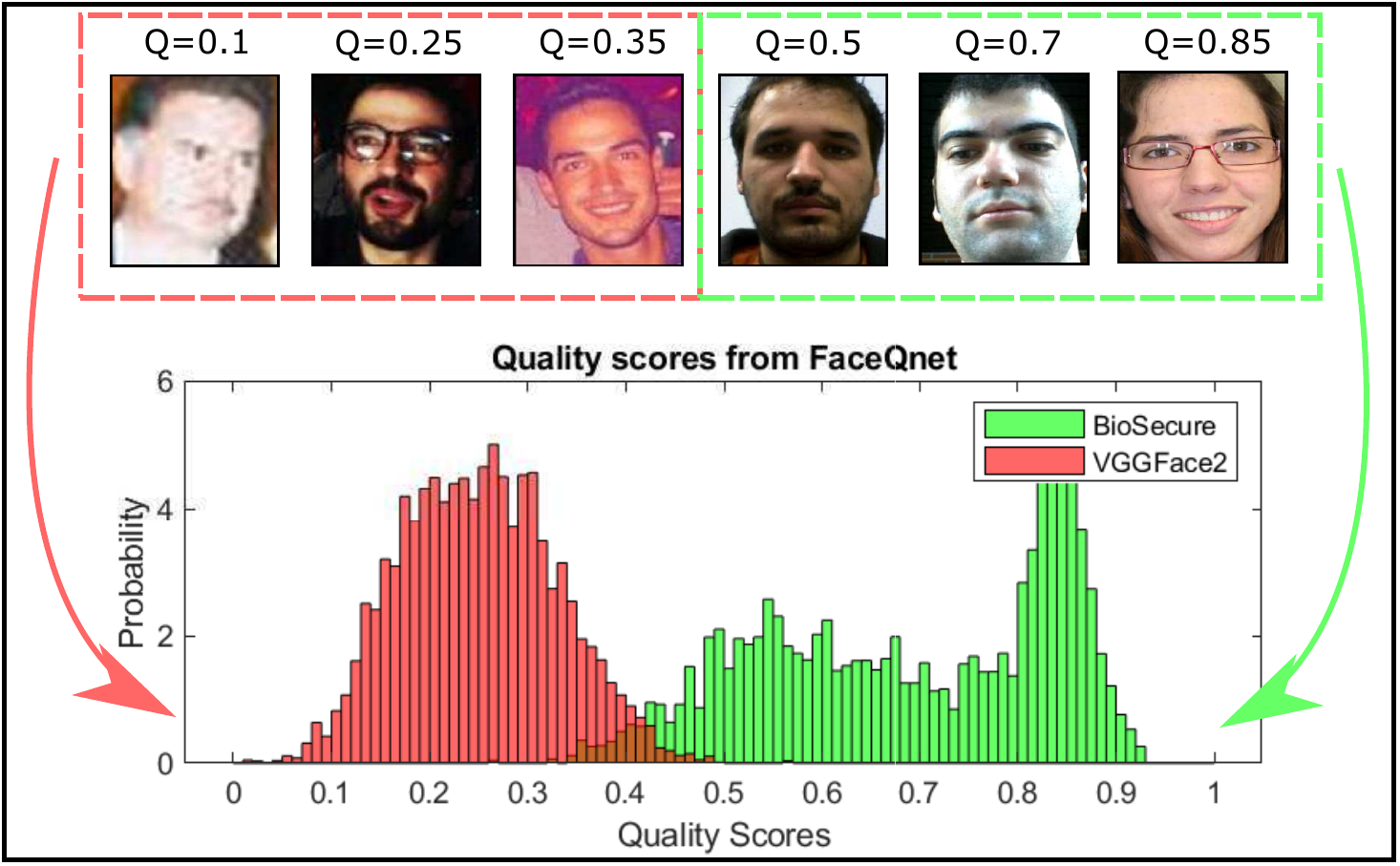}
\end{center}
\caption{\textbf{Distribution of the FaceQnet quality measures for the VGGFace2 and the BioSecure} databases. The images from VGGFace2 obtained lower quality measures compared with those from BioSecure.}
\label{histograma}
\end{figure*}

The recognition accuracy of a specific face comparison depends on the two images begin compared: the gallery and the probe. Comparison scores are therefore dependent on the quality of \textit{both} images. Accordingly, quality assessment is an ill-defined problem: based on only one variable (image $A$) the quality metric has to predict the result of a comparison between two variables (images $A$ and $B$). That is, if a comparison score between two mated images $A$ and B is low, it is not possible to know to what extent this is due to the low quality of $A$, of $B$, or of both. As such, if no constraints are inforced in the problem, a single comparison score cannot be used directly as the quality groundtruth of either image $A$ or $B$.

The question that follows is, how can we generate the groundtruth to train a quality metric based on comparison scores? In the present work we made the assumption that a perfectly compliant ICAO image represents perfect quality. Therefore, if we match such a perfect ICAO image to another image from the same subject and the comparison score is low, it is safe to assume that the second image is of low quality. On the other hand, if the score is high, we can assume that the second image is of good quality. 

This way, by comparing an image $B$ with a perfect ICAO gallery image $A$ of the same subject, we can use the resulting comparison score as the groundtruth quality measure for the image $B$, and then use both image $B$ and its groundtruth quality to train FaceQnet. To achieve this target, we need an ICAO compliance score for each image in the training database. We will use these scores for selecting the gallery images, i.e. the ones with the highest ICAO score. As shown in Figure \ref{general_scheme} (top), to obtain those ICAO compliance scores we used the BioLab framework from \cite{ferrara2012face}. This framework outputs a score between $0$ and $100$ for each individual ICAO compliance test.

As the training set for our quality assessment metric, we selected a subset of $300$ subjects from the VGGFace2 database. For each user we selected as gallery image the one that obtained the highest ICAO compliance score. This way, the training set was divided into gallery images (the one with the highest ICAO compliance score for each subject), and probe images. We decided to employ an existing CNN pretrained for face recognition, the FaceNet model from \cite{schroff2015facenet}, as a feature extractor to get embeddings for all the images in the database. FaceNet was developed in Keras with Tensorflow as its backend. This CNN was trained with the CASIA-WebFace database. It has shown to obtain high accuracy levels in face recognition \cite{schroff2015facenet}. 



First, we used FaceNet to extract $128$-dimensional feature vectors from its last fully-connected layer. Using those embeddings, we computed the similarity between each gallery image and all the other samples of the same user. The similarity values have been normalised to the [$0$,$1$] range. This way a value close to $0$ represents a low quality image (far from the ICAO image), and a value close to $1$ represents a high quality image (close to the ICAO image). The result of the groundtruth generation step described in the present section, was a train dataset used to develop FaceQnet as described in the next section (see Figure~\ref{general_scheme}).


%

\subsection{Regression Model and Training}



In order to build the FaceQnet model, we extended and fine-tuned an existing CNN pretrained for face recognition. The main motivations for this choice were the limited amount of available data, and the relationship between \textit{Quality} and \textit{Accuracy}.

Fine-tuning deep models to perform a task similar to the original one has been successfully used in other studies, where these networks have been used to detect other attributes related to face different than the identity, such as gender, age or race \cite{ranjan2017all}. Since \textit{Quality} and \textit{Accuracy} are closely related, a feature vector that comprises the discriminative information of faces (\textit{Accuracy}), is expected to also comprise the information of their \textit{Quality}.



We selected the ResNet-50 model from \cite{he2016deep}, and removed the classification layer of the pretrained network. We substituted the last layer with two additional Fully Connected (FC) layers to perform regression: one for reducing the dimensionality of the feature vector from $2048$ to $32$ elements, and an ouput layer of size $1$. 

\pagebreak

The first FC layer uses a \textit{ReLu} activation function like the rest of the network, while the output layer does not have any activation. The final architecture of FaceQnet is shown in Figure~\ref{general_scheme} (bottom).

The input to the network are $224\times224\times3$ face images, already cropped and aligned using MTCNN \cite{zhang2016joint}. We froze all the weights of the old layers, and only trained the new ones using the groundtruth quality generated in the previous step (see Section \ref{groundtruth_gen}). 

%
%
%
%
%
%
%

Once trained, FaceQnet can be used as a ``black box" that receives as input a face image, and outputs a quality measure between $0$ and $1$, that is related to the face recognition accuracy. This quality measure can be understood as a proximity measure between the input image and a hypothetical corresponding ICAO compliant sample.

\section{FaceQnet: Evaluation}
\label{evaluation}


The experimental scheme for validating FaceQnet is shown in Figure~\ref{validation_scheme}. In order to test our system, we extracted two test subsets: $100$ subjects from the VGGFace2 database (no overlap with the training set), and $140$ from the BioSecure database. We used FaceQnet to process all the images, obtaining their quality measures, whose distribution can be seen in Figure~\ref{histograma}. The VGGFace2 database presents a higher amount of low quality images since they represent real world acquisition conditions, and therefore, its quality measures are contained in the range [$0$,$0.5$], while the quality values for BioSecure are generally higher (contained in the [$0.3$,$0.95$] range), as its images were acquired in more controlled conditions. We performed face verification comparing each image with all the pictures belonging to the same subject (mated scores). We also compared each image with all the images of another random user (non-mated scores).

We split the images of the VGGFace2 and BioSecure testing datasets into different classes according to their predicted quality measures. We decided to take all the samples from the testing datasets, and we divided them in three ranges: the third of the images with the lowest quality measures (LowQ), the third of the images with medium quality measures (MediumQ), and finally the third with the highest quality scores (HighQ). 

The system used to perform the verification task is a COTS software called Face++ from MEGVII \cite{megviface}. We took the decision of using this API in order to check if our quality metric correlates with the accuracy of a face verification system not used during training. This system performs a comparison between two face images, returning a numerical comparison score between $0$ and $100$. The higher the score, the higher the probability of a mated match. The metrics used to report our results for person verification are the False Acceptance Rate (FAR), the False Rejection Rate (FRR), and the Equal Error Rate (all in \%).

\begin{figure}[t]
\begin{center}
\includegraphics[width=\columnwidth]{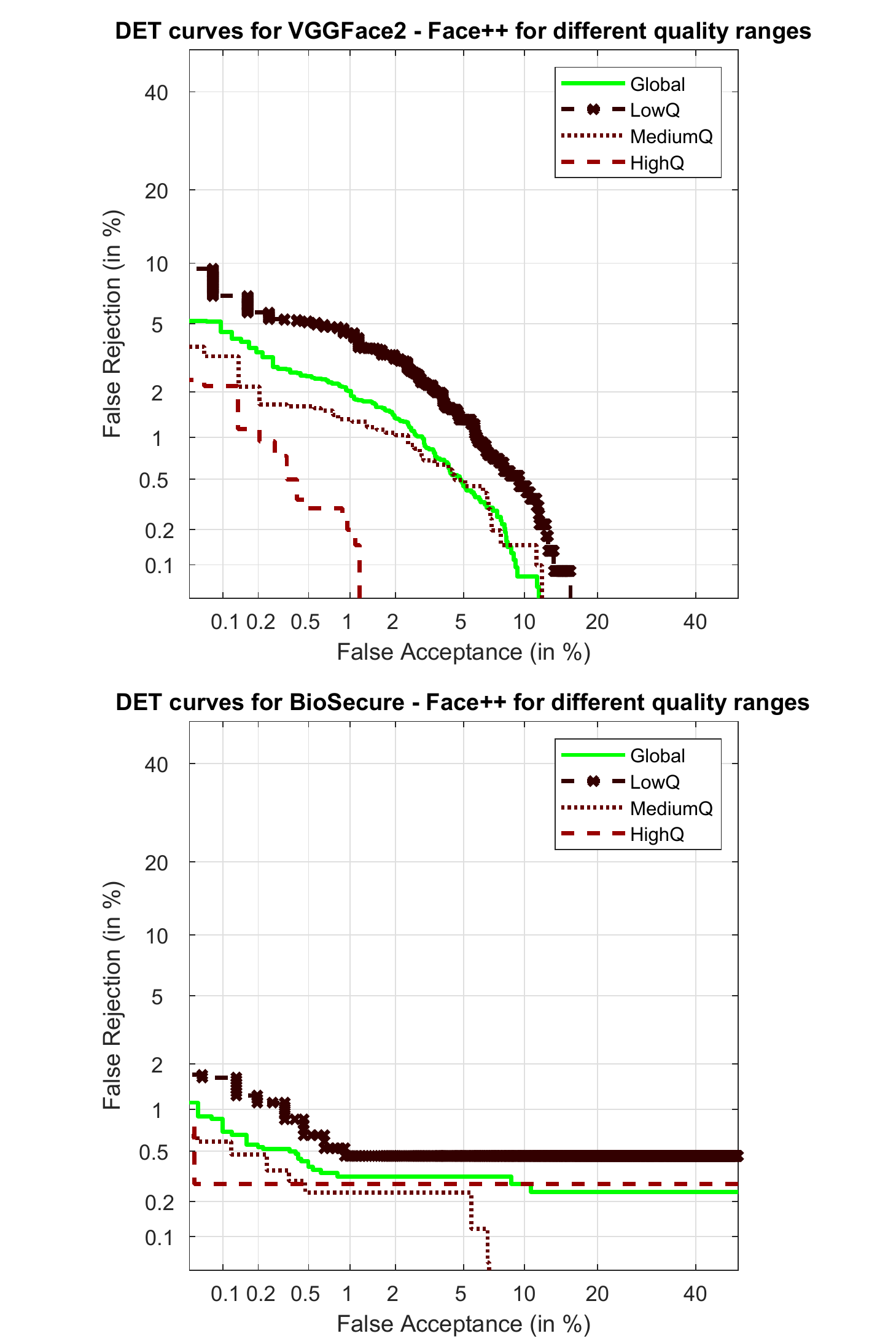}
\end{center}
\caption{\textbf{DET curves} obtained with the Face++ Compare API for the three quality data subsets (LowQ, MediumQ and HighQ) of the VGGFace2 (top) and Biosecure (bottom).}
\label{DETcurves}
\end{figure}

Figure~\ref{DETcurves} shows the DET curves for the different quality ranges. The FAR, FRR, and EER values decrease with the growth of the mean quality of the test samples employed for verification. In general, the performance improves when increasing the mean quality of the samples. The correlation between the quality measure and the verification accuracy is clear even when considering totally different datasets to the one used for training, and a new comparator. 




\section{Conclusion and Future Work}
\label{conclusion_section}

In this paper we studied Quality Assessment (QA) for face recognition. We identified the requirements that a QA tool should have in order to obtain a prediction of the accuracy of a facial image used in face recognition. Our solution consists in employing a CNN previously trained for face recognition, and fine-tuning it with an automatically generated quality groundtruth based on the comparison scores between unconstrained images and perfectly compliant ICAO samples. Using our fine-tuned network, we have extracted quality measures for two subsets of the VGGFace2 and BioSecure databases. These quality measures showed to be a reliable estimation of the face recognition accuracy. We achieved better recognition accuracy using only images of high quality compared to the case in which we used lower quality images. FaceQnet is publicly available in GitHub\footnote{https://github.com/uam-biometrics/FaceQnet}.

As future improvements, a number of different network/algorithms could be used for groundtruth generation, or for obtaining recognition scores, in order to avoid system dependence. Comparison scores produced by different face recognition systems could be fused or combined \cite{2018_INFFUS_MCSreview2_Fierrez} to generate more reliable groundtruth quality values. Further preprocessing of the database could be also helpful for obtaining better results, since the network has shown to be sensitive to outliers in the input data. Additionally, our proposed solution is based on defining quality as an indicator of conformance with ICAO parameters. Other quality definitions or quality groundtruths should be also investigated. In some research topics such as Quality Assessment, the lack of labeled data makes it difficult to train deep neural networks (DNNs) from scratch. To increase the reliability of the predictions, it would be also beneficial to acquire a larger amount of labeled data.

\section{Acknowledgements}
This work was supported in part by projects CogniMetrics from MINECO/FEDER (TEC2015-70627-R), BIBECA from MINECO/FEDER, and BioGuard (Ayudas Fundacion BBVA). The work was conducted in part during a research stay of J. H.-O. at the Joint Research Centre, Ispra, Italy. He is also supported by a PhD Scholarship from UAM.

{\small
\bibliographystyle{ieee}
\bibliography{egbib}
}

\end{document}